\documentclass[runningheads]{llncs}

\usepackage[T1]{fontenc}
\usepackage{graphicx,verbatim}
\usepackage{amsmath,amssymb}
\usepackage{booktabs}
\usepackage{wrapfig}
\usepackage{xcolor}
\usepackage{pifont}
\usepackage{hyperref}
\newcommand{\cmark}{{\color{green!60!black}\ding{51}}}
\newcommand{\xmark}{{\color{red!70!black}\ding{55}}}

\newif\ifusebiblatex
\usebiblatexfalse
\ifusebiblatex
\input{biblatex}
\fi

\begin{document}

\title{MMRareBench:  A Rare-Disease Multimodal and Multi-Image Medical Benchmark}
\titlerunning{MMRareBench}

\author{
  Junzhi Ning\inst{1}\thanks{Equal contribution\quad\textsuperscript{\ding{41}}\,Corresponding author} \and
  Jiashi Lin\inst{1}\textsuperscript{$\star$} \and
  Yingying Fang\inst{2}  \and
  Wei Li\inst{3} \and
  Jiyao Liu\inst{4} \and
  Cheng Tang\inst{1} \and
  Chenglong Ma\inst{4} \and
  Wenhao Tang\inst{1} \and
  Tianbin Li\inst{1} \and
  Ziyan Huang\inst{1} \\
  Guang Yang\inst{2}\textsuperscript{\ding{41}}  
  Junjun He\inst{1}\textsuperscript{\ding{41}}
}

\authorrunning{J. Ning et al.}

\institute{
  Shanghai AI Laboratory \and
  Imperial College London \and
  Shanghai Jiao Tong University \and
  Fudan University\\[3pt]
  \email{hejunjun@pjlab.org.cn}
}

\maketitle

\begin{abstract}
Multimodal large language models (MLLMs) have advanced clinical tasks for common conditions, but their performance on rare diseases remains largely untested. In rare-disease scenarios, clinicians often lack prior clinical knowledge, forcing them to rely strictly on case-level evidence for clinical judgments.
Existing benchmarks predominantly evaluate common-condition, single-image settings, leaving multimodal and multi-image evidence integration under rare-disease data scarcity systematically unevaluated. 
We introduce \textbf{MMRareBench}, to our knowledge the first rare-disease benchmark jointly evaluating multimodal and multi-image clinical capability across four workflow-aligned tracks: diagnosis, treatment planning, cross-image evidence alignment, and examination suggestion. The benchmark comprises 1{,}756 question-answer pairs with 7{,}958 associated medical images curated from PMC case reports, with Orphanet-anchored ontology alignment, track-specific leakage control, evidence-grounded annotations, and a two-level evaluation protocol. A systematic evaluation of 23 MLLMs reveals fragmented capability profiles and universally low treatment-planning performance, with medical-domain models trailing general-purpose MLLMs substantially on multi-image tracks despite competitive diagnostic scores. These patterns are consistent with a \textit{capacity dilution} effect: medical fine-tuning can narrow the diagnostic gap but may erode the compositional multi-image capability that rare-disease evidence integration demands. The benchmark is  available at \href{https://huggingface.co/datasets/junzhin/MMrarebench }{Link}.

\keywords{rare disease \and multimodal \and multi-image \and clinical evaluation \and benchmark}
\end{abstract}

\section{Introduction}



Rare diseases collectively affect over 300 million people worldwide \cite{groft2021progress}. However, each individual condition occurs so infrequently that the resulting long-tail distribution creates severe class imbalance and overlapping phenotypes~\cite{nguengang2020estimating}. 
In this regime, reliable disease-specific priors are sparse, and clinical judgments cannot rely on memorized textbook patterns; they must be strictly grounded in the specific evidence available in the current case. 
To navigate this complexity and reach an accurate diagnosis or treatment decision, clinicians often jointly interpret narrative text, structured signals such as laboratory panels, and multiple medical images spanning modalities and time points. Therefore, successfully diagnosing rare diseases fundamentally depends on two interconnected capabilities: \textbf{(1) multimodal integration}, which requires fusing text, visual, and tabular data to extract complementary signals~\cite{jandoubi2025multimodal}, and \textbf{(2) multi-image analysis}, which demands selecting, aligning, and comparing across several images within one encounter, such as reconciling radiology with pathology or tracking longitudinal change.

Despite the critical need for these capabilities in clinical practice, they are rarely measured together in the development and evaluation of current medical AI. Current multimodal large language models (MLLMs) are  typically assessed on common-condition settings or exam-style, single-image tasks~\cite{lau2018dataset,he2020pathvqa}. 
While some benchmarks focus on rare diseases~\cite{RareBench_KDD2024RareBench}, they primarily evaluate text-based diagnostic accuracy rather than visual evidence integration. Furthermore, many existing settings are textbook-like and internally consistent, which may lead to overestimated performance when clinical evidence is atypical or partially contradictory~\cite{ye2024gmai,pal2022medmcqa,jin2019pubmedqa}. 

\begin{table}[t]
    \centering
    \caption{\textbf{Benchmark comparison.} Rare: rare-disease focus; M.mod: multi-modality inputs (text, image, table); M.img: multi-image per case (multiple images); Clinical: workflow-aligned clinical tracks.}
    \label{tab:benchmark_comparison}
    \footnotesize
    \renewcommand{\arraystretch}{1}
    \begin{tabular}{@{}lcccc@{}}
        \toprule
        \textbf{Benchmark} & \textbf{Rare} & \textbf{M.mod} & \textbf{M.img} & \textbf{Clinical} \\
        \midrule
        MedR-Bench~\cite{MedRBench_2025}   & \xmark & \xmark & \xmark & \cmark \\
        MTBBench~\cite{MTBBench_NeurIPS2025} & \xmark & \cmark & \cmark & \cmark \\
        RareBench~\cite{RareBench_KDD2024RareBench} & \cmark & \xmark & \xmark & \xmark \\
        GMAI-MMBench~\cite{ye2024gmai}         & \xmark & \cmark & \xmark & \cmark \\
        \textbf{Ours}                        & \cmark & \cmark & \cmark & \cmark \\
        \bottomrule
    \end{tabular}
\vspace{-20pt}
\end{table}

To address these gaps, we present \textbf{MMRareBench}, a multimodal rare-disease benchmark that stress-tests whether MLLMs truly integrate complex case evidence under rare-disease data scarcity, rather than rely on memorized common-condition patterns.
Constructed from thousands of real-world clinical case reports and mapped to the Orphanet rare-disease nomenclature, MMRareBench ensures medical authenticity while enabling standardized diagnosis normalization. Because case-report-based evaluation can be inflated by shortcut cues when answers leak into the context, we build leakage-resistant masked views with per-item \texttt{leakage\_audit} records to support auditing and evidence-grounded assessment.
The benchmark spans four workflow-aligned clinical tracks: Diagnosis track T1, Treatment Planning track T2, Cross-Image Evidence Alignment track T3, and Examination Suggestion track T4. Performance is measured using a two-level protocol that combines deterministic verification with model-graded rubric scoring. Table~\ref{tab:benchmark_comparison} summarizes key differences between MMRareBench and existing works.

\textbf{Contributions:} 
(1) We introduce MMRareBench, to our knowledge the first Orphanet-aligned multimodal, multi-image benchmark for rare diseases, constructed from real-world clinical case reports.
(2) We define a workflow-aligned evaluation tasks spanning four tracks, with structured targets and track-specific rubrics under rare-disease data scarcity.
(3) We benchmark 23 MLLMs and uncover systematic capability gaps under the rare-disease workflow, including a non-trivial treatment-planning bottleneck and a \textit{capacity dilution} pattern where medical fine-tuning narrows diagnosis error yet degrades cross-image evidence alignment.

\section{Related Work}

\textbf{Medical benchmarks.} Early medical benchmarks focus on text-only, exam-style QA \cite{MedQA_Jin_2021}, but lack grounding in clinical workflows. Recent work introduces structured evaluation via case reports, with MedR-Bench \cite{MedRBench_2025} modeling multi-stage workflows and MedCaseReasoning \cite{wu2025medcasereasoning} emphasizing rationale-based diagnostic inference. Multimodal extensions further expand this landscape: GMAI-MMBench \cite{ye2024gmai} provides a comprehensive evaluation for general medical AI, while MTBBench \cite{MTBBench_NeurIPS2025} targets sequential clinical decision-making, and MedFrameQA \cite{yu2025medframeqa} explores multi-image VQA. However, none of these benchmarks jointly evaluates multimodal and multi-image evidence integration under the rare-disease long tail.

\textbf{Rare-disease benchmarks.} Rare-disease evaluation is complicated by long-tail distributions, phenotypic overlap, and limited per-condition data. Existing benchmarks are predominantly text-driven and diagnosis-oriented. RareBench~\cite{RareBench_KDD2024RareBench} formulates phenotype-to-code mapping, while ReDis-QA~\cite{wang2024assessing} frames rare-disease diagnosis as question answering over structured and unstructured clinical descriptions. 
CaseReportBench~\cite{zhang2025casereportcollective} emphasizes narrative understanding and diagnostic inference over real-world rare case reports. NOVA~\cite{bercea2025nova} introduces multimodal MRI anomaly detection, but focuses on lesion localization rather than end-to-end diagnostic workflows. None of these benchmarks jointly evaluates multimodal and multi-image evidence integration under the data-scarce rare-disease long tail. MMRareBench fills this gap with workflow-aligned tracks anchored to ORPHAcodes.

\section{Method}

\subsection{Data Source and Curation Pipeline}
We collect 14,700 PMC case reports published between 2005 and 2025, using the PMCID list provided by \cite{wu2025medcasereasoning}.
These reports were processed into 1,756 benchmark items covering 403 rare diseases through a five-stage pipeline.
Rare-disease case reports pose distinctive curation challenges because long-tail distributions and overlapping phenotypes demand ontology-anchored filtering that generic clinical NLP pipelines cannot provide. Each report is parsed into typed document blocks, each assigned a persistent identifier that serves as an evidence anchor throughout the pipeline. An LLM-assisted extraction step then identifies imaging modalities, disease mentions, and clinical summaries from these blocks. Disease candidates are matched to the Orphanet rare-disease nomenclature; only cases whose primary diagnosis maps to a confirmed ORPHA code are retained. Documents are then assigned to track-specific buckets, retaining only single-patient reports.

\begin{figure}[t]
    \centering
    \includegraphics[width=\textwidth]{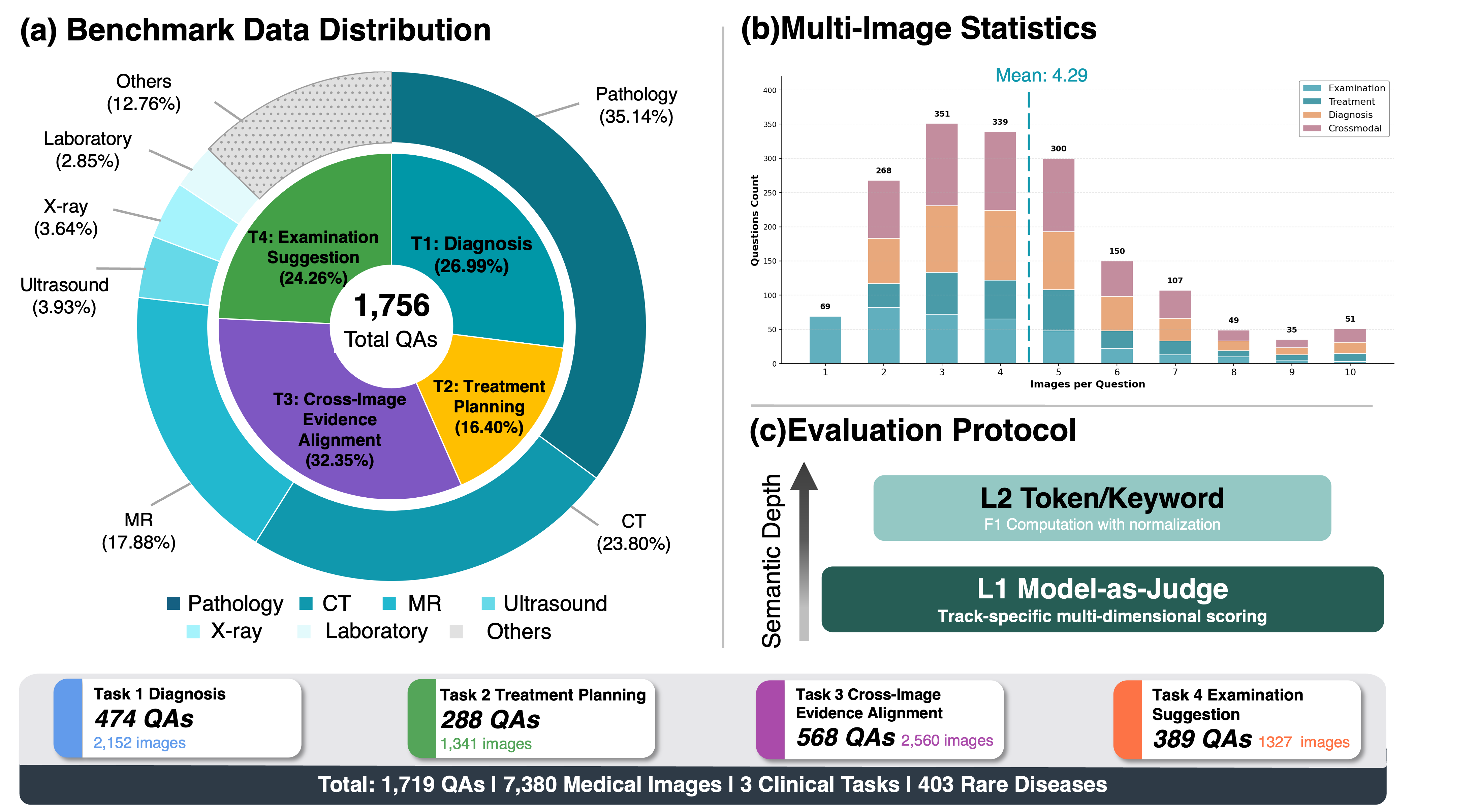}
    \vspace{-6pt}
    \caption{\textbf{MMRareBench overview.}
    \textbf{(a)}~Track and modality distribution across 1,756 items.
    \textbf{(b)}~Images per sample (mean\,=\,4.5).
    \textbf{(c)}~Two-level evaluation protocol: L1 model-graded rubric scoring and L2 token-level F1.}
    \vspace{-20pt}
    \label{fig:benchmark_stats}
\end{figure}

\subsection{Leakage Control and QA Generation}
\label{sec:composer}

Clinical benchmarks for rare diseases risk becoming trivially solvable when diagnostic conclusions leak into the input context. To enforce genuine, evidence-grounded reasoning, we construct track-specific masked views implemented via three cascading layers. \textbf{Layer~1} removes metadata and summary sections (title, abstract, conclusion) across all tracks to isolate the clinical narrative. \textbf{Layer~2} replaces the primary diagnosis and its Orphanet aliases with \texttt{[DIAGNOSIS\_MASKED]} for T1 and T4. \textbf{Layer~3} redacts figure and table captions with \texttt{[CAPTION\_MASKED]} for T1, T3, and T4. Each item carries a \texttt{leakage\_audit} record, and a strict pre-check rejects any sample where the standardized diagnosis circumvents the mask. These masked views and associated images are then submitted to GPT-5.1~\cite{singh2025openai} to generate a track-specific question, a reference answer, and structured targets (e.g., plan elements for T2, relation types for T3, key tests for T4). Crucially, the model must ground its outputs by generating an evidence chain of two to five claims, each citing a specific evidence-unit identifier with a supporting quote. Finally, candidate items undergo automated auditing for schema validity, image integrity, and residual answer-in-context leakage. This automated filtering removes approximately 30\% of candidates, and the retained items proceed to stratified human verification.

\subsection{Task Definition: Four Clinical Tracks}

\begin{figure}[t]
\centering
\includegraphics[width=\textwidth]{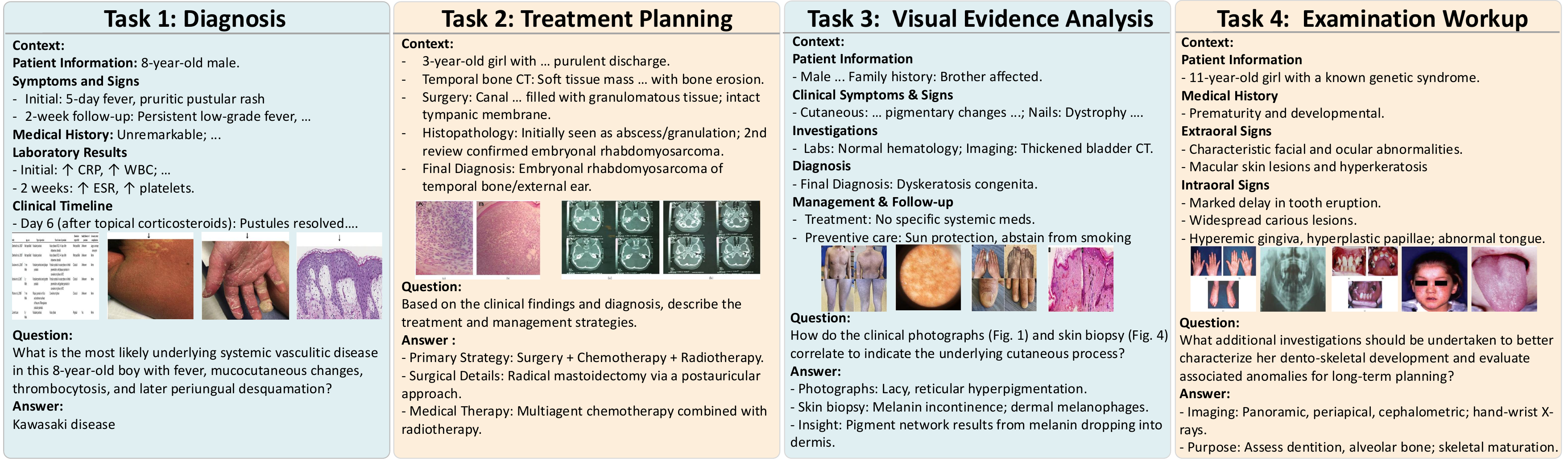}
\vspace{-6pt}
\caption{\textbf{Representative examples from the four clinical tracks.} Note that to illustrate the core structure, we present a simplified version of the demo in this case; however, full details are comprehensively outlined in the benchmark.}
\vspace{-5pt}
\label{fig:task_examples}
\end{figure}

MMRareBench defines four tracks aligned with key decision points in a rare-disease diagnostic workflow, where phenotypic overlap and atypical presentations make each stage substantially harder than in common conditions. Across all tracks, the model receives a de-identified clinical narrative together with multiple medical images, and track-specific masked views described in Sec.~\ref{sec:composer} remove direct label leakage while preserving clinical realism.  Fig.~\ref{fig:benchmark_stats} visualizes track and modality composition alongside the evaluation hierarchy, and Fig.~\ref{fig:task_examples} shows representative examples from each track.

\textbf{T1 Diagnosis} requires predicting the primary diagnosis in open-ended form from a masked case narrative where explicit diagnosis mentions have been removed, together with the associated medical images. This track tests hypothesis formation under information scarcity in rare diseases, where phenotypic overlap is common and imaging findings can be atypical.

\textbf{T2 Treatment Planning} provides the working diagnosis, the full case narrative, and associated medical images, and asks the model to generate a stage-wise plan emphasizing safety, deferral under uncertainty, monitoring, and follow-up. This track evaluates whether a model can translate a diagnostic hypothesis into actionable clinical decisions when standardized protocols for rare conditions are limited or absent.

\textbf{T3 Cross-Image Evidence Alignment} is the track that explicitly isolates multi-image, cross-modality evidence integration. Given at least two medical images from different imaging modalities with the accompanying narrative, the model must describe salient findings per image, explicitly align evidence across images by identifying how findings relate and predict a relation type. This capability is clinically central in rare diseases where imaging evidence is often partial, atypical, or internally inconsistent, and cross-modality correlation is essential for disambiguation.

\textbf{T4 Examination Suggestion} presents a masked case narrative truncated at a temporal cut-point together with the associated medical images, and asks the model to recommend prioritized next-step investigations with rationale; gold-standard tests are withheld as the expected answer. This track tests whether the model can propose discriminative, high-yield workups to reduce remaining diagnostic uncertainty for rare conditions rather than restating observed evidence.

\subsection{Evaluation Metrics}


As illustrated in Fig.~\ref{fig:benchmark_stats}(c), we adopt a two-level evaluation protocol that combines (L1) model-graded rubric scoring and (L2) deterministic verification for semantic and clinical adequacy.


\paragraph{\textbf{Model-graded rubric scoring.}}
We use Qwen3-VL-235B~\cite{yang2025qwen3} as the judge with track-specific rubric prompts. Each prompt supplies the reference answer, a set of per-dimension evaluation criteria, and a strict-calibration instruction that defaults to \emph{no} and awards credit only when a criterion is fully satisfied. Dimension counts and aggregation differ by track: T1 uses 3 weighted binary criteria; T2 and T3 average 6 and 5 binary YES/NO ratings uniformly; T4 aggregates 8 ordinal ratings on a 0--2 scale, normalized by 16. Scores are averaged over the dataset. Prompts are validated on a human-annotated subset.

\paragraph{\textbf{Track-specific rubric design.}}
All rubrics enforce a strict-calibration principle to counter the leniency bias of LLM-as-judge pipelines: the judge defaults to NO and awards credit only when a criterion is completely satisfied.

\textbf{T1} employs a weighted cascade over three binary dimensions, namely exact diagnosis match, disease category, and diagnostic specificity, where exact match carries three times the weight of the other two. When exact match fails, the total score is capped at~0.2.

\textbf{T2} and \textbf{T3} apply uniform binary ratings across six and five dimensions respectively. T2 targets treatment specificity, drug-level accuracy, safety, follow-up concreteness, pathophysiological justification, and clinical implementability. T3 progresses from per-modality finding identification through cross-modal alignment to integrated synthesis, with the final dimension testing whether the response derives new clinical insight from combining modalities.

\textbf{T4} uses an ordinal 0--2 scale across eight dimensions covering test coverage, specificity, conditional ordering, expected findings, quantitative criteria, differential narrowing, patient-specific adaptation, and testing precision. Scores are normalized by 16, with calibrated anchors that reserve a score of~2 for textbook-level performance and assign 1 for competent responses.

\paragraph{\textbf{Deterministic verification (L2).}}
For T1, we also apply SQuAD-style normalization~\cite{rajpurkar2016squad} and compute token-level F1 between the predicted answer $y_i$ and gold answer $y_i^*$, taking the maximum over aliases:
\begin{equation}
F_1 = \frac{1}{N} \sum_{i=1}^{N} \frac{2 \cdot |\text{tokens}(y_i) \cap \text{tokens}(y_i^*)|}{|\text{tokens}(y_i)| + |\text{tokens}(y_i^*)|}
\end{equation}

\section{Experiments and Analysis}

We benchmark 23 multimodal large language models on MMRareBench, grouped into closed-source models in Table~\ref{tab:closed_results} and open-source models in Table~\ref{tab:open_results}, including 9 medical-domain models, to systematically address three core research questions: \textbf{(1)}~\textit{Performance landscape.} How do current MLLMs perform across the four clinical tracks, and which stages of the rare-disease workflow present the greatest challenge? \textbf{(2)}~\textit{Domain specialization gap.} Does medical-domain fine-tuning improve rare-disease clinical performance, and how do specialized medical models compare with general-purpose MLLMs? \textbf{(3)}~\textit{Evaluation quality beyond headline scores.} Do models that achieve high task scores also demonstrate consistent cross-track performance?

\subsection{Overall Performance Landscape (RQ1)}

\paragraph{\textbf{Treatment Planning is the hardest track for rare-disease MLLMs}}
Closed-source MLLMs achieve the highest scores on every track, yet T2 presents the most severe performance bottleneck across all model categories as reported in Table~\ref{tab:closed_results} and Table~\ref{tab:open_results}, with even the best model achieving a score of only 49.2. Treatment planning compounds multiple sources of difficulty: it presupposes accurate diagnosis, requires integration of medical imaging evidence, and demands risk-aware judgment about interventions for which standard protocols are largely absent from training corpora.

\begin{table}[!htbp]
\centering
\renewcommand{\arraystretch}{0.85}
\setlength{\aboverulesep}{0.3ex}
\setlength{\belowrulesep}{0.3ex}
\caption{Performance of closed-source MLLMs on MMRareBench. We report model-graded scores for all tracks (Score; 0--100), with token-level F1 reported for diagnosis (T1). \textbf{Bold}: best; \underline{underline}: second best.}
\label{tab:closed_results}
\resizebox{0.9\textwidth}{!}{%
\begin{tabular}{l cc c c c}
\toprule
\toprule

 & \multicolumn{2}{c}{\textbf{T1 Diag.}} & \multicolumn{1}{c}{\textbf{T2 Treat.}} & \multicolumn{1}{c}{\textbf{T3 Cross.}} & \multicolumn{1}{c}{\textbf{T4 Exam.}} \\
\cmidrule(lr){2-3} \cmidrule(lr){4-4} \cmidrule(lr){5-5} \cmidrule(lr){6-6}
\textbf{Model} & Score & F1 & Score & Score & Score \\
\midrule
GPT-5~\cite{singh2025openai} & \underline{75.9} & 76.1 & \textbf{49.2} & \textbf{70.8} & 77.9 \\
GPT-4o~\cite{hurst2024gpt} & 62.5 & 70.0 & 12.7 & 15.6 & 42.8 \\
Claude-Haiku-4.5~\cite{anthropic2025claude45haiku} & 30.9 & 51.8 & 34.2 & 41.9 & 80.6 \\
Claude-Sonnet-4.5~\cite{anthropic2025claude45sonnet} & 68.5 & 70.8 & \underline{35.8} & 54.7 & \textbf{85.9} \\
Gemini-2.5-Flash~\cite{comanici2025gemini} & 62.2 & 70.0 & 17.1 & 54.9 & 66.9 \\
Gemini-2.5-Pro~\cite{comanici2025gemini} & 74.0 & \underline{80.1} & 26.7 & \underline{66.9} & \underline{85.5} \\
Gemini-3-Flash-Preview~\cite{google2025gemini3} & \textbf{76.6} & \textbf{84.7} & 29.1 & 62.9 & 81.8 \\
\bottomrule
\bottomrule

\end{tabular}}
\vspace{-5pt}
\end{table}
\begin{table}[t]
\centering
\renewcommand{\arraystretch}{0.7}
\setlength{\aboverulesep}{0.3ex}
\setlength{\belowrulesep}{0.3ex}
\caption{Performance of open-source MLLMs on MMRareBench. Med.: medical-domain model. Same metrics as Table~\ref{tab:closed_results}.}
\label{tab:open_results}
\resizebox{\textwidth}{!}{%
\begin{tabular}{l c cc c c c}
\toprule
\toprule

 &  & \multicolumn{2}{c}{\textbf{T1 Diag.}} & \multicolumn{1}{c}{\textbf{T2 Treat.}} & \multicolumn{1}{c}{\textbf{T3 Cross.}} & \multicolumn{1}{c}{\textbf{T4 Exam.}} \\
\cmidrule(lr){3-4} \cmidrule(lr){5-5} \cmidrule(lr){6-6} \cmidrule(lr){7-7}
\textbf{Model} & \textbf{Med.} & Score & F1 & Score & Score & Score \\
\midrule
\multicolumn{7}{l}{\textit{General-purpose}} \\
\midrule
GLM-4.6v~\cite{hong2025glm} &  & 25.8 & 30.1 & \underline{20.5} & 28.7 & 18.7 \\
Qwen3-VL-235B-Instruct~\cite{yang2025qwen3} &  & \textbf{58.1} & \textbf{67.3} & \textbf{35.9} & \textbf{56.6} & \textbf{83.2} \\
Qwen3-VL-30B-Instruct~\cite{yang2025qwen3} &  & 46.9 & 56.3 & 20.1 & \underline{36.9} & \underline{63.1} \\
Qwen3-VL-8B-Instruct~\cite{yang2025qwen3} &  & 39.3 & 46.7 & 10.8 & 25.7 & 43.9 \\
Qwen2.5-VL-72B~\cite{ahmed2025qwen} &  & \underline{52.6} & \underline{61.0} & 8.7 & 18.9 & 42.3 \\
Qwen2.5-VL-32B~\cite{ahmed2025qwen} &  & 46.9 & 49.6 & 12.8 & 21.2 & 42.0 \\
Qwen2.5-VL-7B~\cite{ahmed2025qwen} &  & 33.7 & 41.0 & 7.2 & 5.5 & 30.0 \\
\midrule
\multicolumn{7}{l}{\textit{Medical-domain}} \\
\midrule
Hulu-Med-14B~\cite{jiang2025hulu} & $\checkmark$ & 52.0 & 58.2 & 14.2 & 6.1 & 38.8 \\
Hulu-Med-7B~\cite{jiang2025hulu} & $\checkmark$ & 42.2 & 49.0 & 11.3 & 4.9 & 34.1 \\
Hulu-Med-4B~\cite{jiang2025hulu} & $\checkmark$ & 39.8 & 45.1 & 10.4 & 4.3 & 37.2 \\
Lingshu-32B~\cite{xu2025lingshu} & $\checkmark$ & 49.8 & 53.3 & 14.2 & 8.7 & 36.9 \\
Lingshu-7B~\cite{xu2025lingshu} & $\checkmark$ & 37.9 & 42.9 & 12.1 & 4.0 & 35.1 \\
Medgemma-27b-it~\cite{sellergren2025medgemma} & $\checkmark$ & 43.7 & 51.3 & 15.4 & 27.3 & 56.0 \\
Medgemma-4b-it~\cite{sellergren2025medgemma} & $\checkmark$ & 30.5 & 38.0 & 5.5 & 2.9 & 33.1 \\
\midrule
\multicolumn{7}{l}{\textit{Medical unified understanding \& generation}} \\
\midrule
HealthGPT-L14-Compre~\cite{lin2025healthgpt} & $\checkmark$ & 36.0 & 41.8 & 14.1 & 0.6 & 20.4 \\
UniMedVL~\cite{ning2025unimedvl} & $\checkmark$ & 18.3 & 8.1 & 9.6 & 0.9 & 32.5 \\
\bottomrule
\bottomrule

\end{tabular}}
\vspace{-10pt}
\end{table}

\subsection{Rare Disease Domain Specialization Gap (RQ2)}

\begin{figure}[t]
\centering
\includegraphics[width=0.95\textwidth]{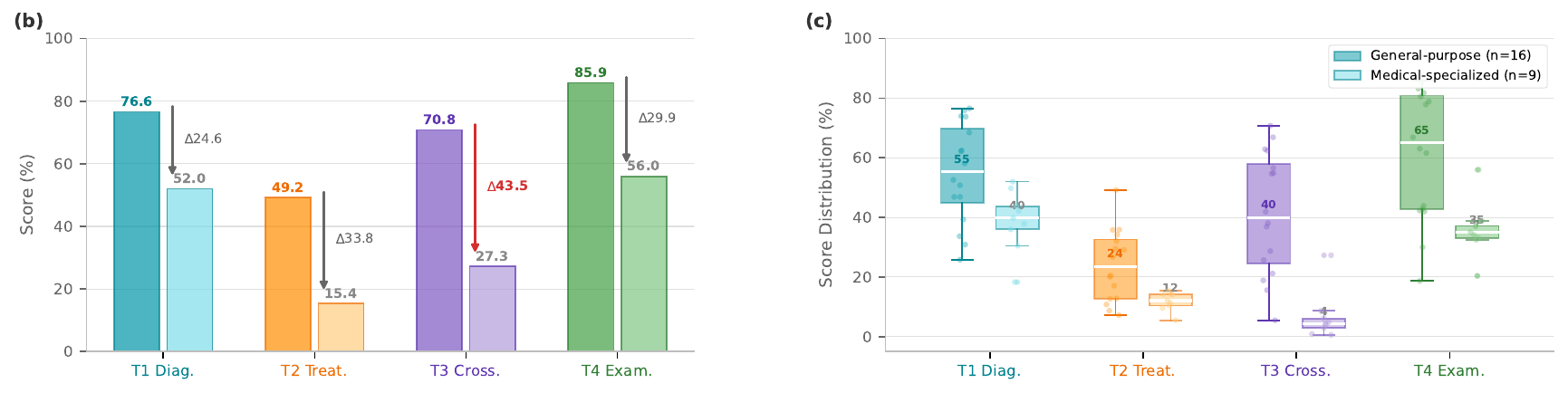}
\vspace{-6pt}
\caption{\textbf{Domain specialization gap.}
Metrics per track: Score (T1--T4).
\textbf{(a)}~Peak general-purpose vs.\ medical scores per track;
\textbf{(b)}~Score distributions for general-purpose  and medical models.}
\vspace{-10pt}
\label{fig:results}
\end{figure}

\paragraph{\textbf{Medical-domain MLLMs close the Diagnosis gap but trail on Cross-Image Evidence Alignment}}
On T1, the best medical MLLM trails the best general-purpose open-source model by only 6 points, as diagnosis most closely resembles the pattern-recognition tasks targeted by medical fine-tuning. However, the gap widens sharply on multi-image tracks: on T3, the best medical model trails the best general-purpose open-source model by over 29 points and the best closed-source model by 43.5 points, as shown in Fig.~\ref{fig:results}a,b. These results expose a fundamental \textit{capacity dilution} effect: medical fine-tuning on limited-parameter models may concentrate capacity on common-condition pattern recall at the expense of compositional multi-image evidence integration, while general-purpose models retain transferable capabilities that prove decisive.

\paragraph{\textbf{Within medical MLLM families, model scale improves all tracks but does not close the Cross-Image evidence gap}}
Within each medical MLLM family, larger variants consistently outperform smaller ones across all tracks, as Table~\ref{tab:open_results} confirms. However, even the largest medical model retains a substantial deficit on T3 relative to the best closed-source model, confirming that cross-image evidence alignment rather than raw parameter count is the binding constraint for rare-disease clinical evaluation.

\subsection{Evaluation Quality Beyond Headline Scores (RQ3)}

\paragraph{\textbf{Single-track gains in MLLMs do not yield consistent performance across the rare-disease workflow.}}
Individual MLLM strengths do not transfer across tracks. For example, GPT-4o achieves 62.5 on T1 but only 15.6 on T3, highlighting cross-image evidence alignment as the most discriminating capability dimension. As Table~\ref{tab:closed_results} shows, no evaluated model ranks first on more than two tracks simultaneously, indicating that rare-disease clinical evaluation requires a combination of capabilities that no current architecture fully integrates.

\section{Conclusion}

Under rare-disease long-tail distributions, disease-specific priors are sparse and unreliable, making multimodal and multi-image evidence the primary rather than supplementary information source for clinical judgment. MMRareBench addresses this critical gap. Benchmarking 23 MLLMs reveals fragmented capability profiles, universally low treatment-planning performance, and patterns consistent with \textit{capacity dilution}: medical fine-tuning can narrow the diagnostic gap but may erode cross-image evidence alignment by concentrating capacity on common-condition pattern recall. MMRareBench's track-structured evaluation provides a reproducible testbed for progress toward evidence-grounded multimodal clinical evaluation in rare diseases.

\ifusebiblatex
\PrintBibliography
\else
\bibliographystyle{splncs04}
\bibliography{references}
\fi

\end{document}